\documentclass[11pt]{article}

\usepackage{arxiv}

\usepackage{graphicx}
\usepackage[title]{appendix}
\usepackage{xcolor}
\usepackage{amssymb}
\usepackage{amsmath}
\usepackage[section]{placeins}
\usepackage{subcaption}
\usepackage{booktabs}
\usepackage{algorithm}
\usepackage{algpseudocode}
\usepackage{newtxtext,newtxmath}
\usepackage{bm}
\usepackage{natbib}
\usepackage[colorinlistoftodos,prependcaption,textsize=tiny]{todonotes}

\usepackage{array}
\usepackage{arydshln}
\DeclareMathOperator*{\argminA}{arg\,min} 
\newcommand{\floor}[1]{\left\lfloor #1 \right\rfloor}

\title{DynNet: Physics-based neural architecture design for linear and nonlinear structural response modeling and prediction}

\author{
  Soheil Sadeghi Eshkevari\thanks{ses516@lehigh.edu} \\
  Lehigh University\\
  \texttt{ses516@lehigh.edu} \\
  \And
  Martin Tak\'a\v{c}\thanks{mat614@lehigh.edu} \\
  Lehigh University \\
  \texttt{mat614@lehigh.edu} \\
  \And
  Shamim N. Pakzad\thanks{snp208@lehigh.edu} \\
  Lehigh University\\
  \texttt{pakzad@lehigh.edu} \\
  \And
  Majid Jahani\thanks{maj316@lehigh.edu} \\
  Lehigh University \\
  \texttt{maj316@lehigh.edu} \\
  }

\begin{document}
\maketitle

\begin{abstract}

Data-driven models for predicting dynamic responses of linear and nonlinear systems are of great importance due to their wide application from probabilistic analysis to inverse problems such as system identification and damage diagnosis. In this study, a physics-based recurrent neural network model is designed that is able to learn the dynamics of linear and nonlinear multiple degrees of freedom systems given a ground motion. The model is able to estimate a complete set of responses, including displacement, velocity, acceleration, and internal forces. Compared to the most advanced counterparts, this model requires smaller number of trainable variables while the accuracy of predictions is higher for long trajectories. In addition, the architecture of the recurrent block is inspired by differential equation solver algorithms and it is expected that this approach yields more generalized solutions. In the training phase, we propose multiple novel techniques to dramatically accelerate the learning process using smaller datasets, such as hardsampling, utilization of trajectory loss function, and implementation of a trust-region approach. Numerical case studies are conducted to examine the strength of the network to learn different nonlinear behaviors. It is shown that the network is able to capture different nonlinear behaviors of dynamic systems with very high accuracy and with no need for prior information or very large datasets. 

\end{abstract}

\section{Introduction}
\label{sec:intro}

Dynamic response prediction of structural systems has been a great tool for design and assessment of individual buildings as well as reliability analysis of infrastructure and large urban areas. Traditionally, this process is executed by building numerical models of dynamic systems and predicting responses using numerical differential equation solvers such as Newmark-$\beta$ method. However, this approach is suitable for structures with known physical properties (i.e., mass, stiffness, and damping matrices) with very accurate analytical modals for nonlinear components of the structures. Structural health monitoring (SHM) methods have been effective in identifying mechanical properties of the existing structures. Yet, the dynamic response simulation of an existing system requires a comprehensive SHM phase for model updating \citep{yuen2006efficient,ching2004new,johnson2004phase,shahidi2014generalized}. In addition, for an accurate simulation of a structure with nonlinear components, emerging technologies such as real-time hybrid simulation are proposed \citep{christenson2008large,ahmadizadeh2008compensation,al2020assessment}. This approach is also limited to individual nonlinear structural components and requires advanced experimental and numerical devices. \par

Artificial intelligence has been one of the most useful and promising tools in science and technology over the past few decades. In particular, machine learning has demonstrated a great potential for learning and predicting nonlinear behaviors and trends in large and noisy datasets \citep{deng2014deep}. Neural Networks (NN) have shown an exceptional potential as universal function approximators with minimal need for prior information about the underlying knowledge of a problem \citep{cybenko1989approximation,leshno1993multilayer}. However, in engineering applications, black-box function approximators are less favored due to the fact that for many of those, solid underlying equations/models exist. Knowledge-based machine learning approach intends to bridge this gap by contributing governing equations into machine learning models \citep{towell1990refinement}.

\subsection{Artificial Intelligence in Civil Engineering}

In general, the major applications of machine learning in civil engineering can be divided into following categories: (a) system identification (SID); (b) damage detection; and (c) dynamic response prediction of structural systems. A detailed overview of machine learning algorithms for damage detection is given in \citet{worden2007application} and \citet{ying2013toward}. In summary, the methods use machine learning algorithms (e.g., support vector machines (SVM) and multi-layer perceptrons (MLP)) for classification between damaged and undamaged states of structural components based on low-level inputs (e.g., motion sensor data). A multi-stage damage detection method is proposed in \citep{yi2013multi} in which signal features are extracted using wavelet transforms and an MLP network diagnoses whether damage has occurred. \citet{gui2017data} proposed a method for feature extraction from sensor data time series and damage classification based on these extracted features using SVM. More recently, end-to-end damage detection algorithms are emerging in which feature extraction and damage detection stages are combined in a single estimator. \citet{abdeljaber2017real} proposed a vibration-based convolutional neural network (CNN) for direct damage detection and localization based on sensor time signals. \citet{gulgec2019convolutional} proposed a one-step vision-based damage detection and localization method via CNN which uses 2D strain fields as input. \par

Fewer studies have investigated data-driven methods for system identification due to the inherited model-dependency of this problem. Some efforts have been made to reconstruct underlying equations using data-driven algorithms. \citet{brunton2016discovering} proposed a look-up approach to reconstruct the governing equation of dynamic systems using sparse identification. More recent studies investigate machine learning solutions with model-guided constraints. \citet{raissi2018hidden} introduced hidden physics models that are able to identify underlying physics of dynamic systems using small datasets. In civil engineering, \citet{sadeghi2020modal} proposed a data-driven approach for bridge modal identification using mobile sensing data. The model is highly constrained by the modal superposition law of structural dynamics and could successfully identify complete modal properties. \par

In addition to diagnosis and monitoring tasks that are objectives of the previous studies, data-driven approaches for dynamic response prediction of structural systems has been of great importance and interest. Finite element analysis (FEA) along with nonlinear time history analysis (NTHA) has enabled very accurate dynamic response estimations; however, both techniques are computationally expensive and require detailed information of the system. By emergence of probabilistic reliability analyses of individual and clusters of structures subject to hazards (e.g., earthquake), it is realistically impractical to carry out extensive FEA and NTHA analyses of increasingly larger systems \citep{song2010multi,mahsuli2013seismic}. Therefore, faster, more flexible, and reliable approaches are highly required. 


\subsection{Data-driven Dynamic Response Prediction}

Dynamic response prediction of structures using statistical methods have been widely investigated over the last few years. The approaches span from model-based predictions to data-driven models such as autoregressive moving average (ARMA) models or neural networks. A model-based full state predictor is proposed that incorporates a prior nonlinear model of the building for experimental response prediction \citep{roohi2019nonlinear}. \citet{mattson2006statistical} proposed an autoregressive model to predict major trends of the dynamic response; however, the effect of exogenous input was remained and considered as residual. In fact, despite their simplicity, ARMA-based models are limited to stationary and linear systems. To address that, \citet{bornn2009structural} proposed an autoregressive SVM that incorporates nonlinear functionalities within the prediction equation. Neural networks (NN) have been the most recent approach for dynamic response prediction due to their flexibility and great performance in regression problems. The pioneer studies were focused on simple MLP models for partial one-step ahead response prediction (i.e., predictions include some but not all of the followings: displacement, velocity, acceleration, and internal force at all degrees of freedom). \citet{lightbody1996multi} proposed a single layer neural network in which the output is a weighted sum of multiple trainable AR models with \textit{Tanh} activation. The study was a breakthrough that enhanced estimator complexities from individual linear model to a nonlinear ensemble of linear models. By recent computational developments, deeper MLP networks were utilized for more comprehensive dynamic response predictions of nonlinear cases. \citet{lagaros2012neural} proposed a MLP for one-step ahead response prediction of nonlinear buildings. The method showed great performance both numerically and experimentally, however, the prediction was limited to displacement time histories. Note that in general there is no guarantee for reasonable predictions of other response components (e.g., velocity and accelerations) using a single component when using data-driven regression methods. Therefore, yet more comprehensive predictive models are required. \par

Theoretically speaking, MLPs are ideal when the input features are fully independent. In dynamic response prediction problem, however, a high inter-dependency between responses at consequent time steps exist. Therefore, other neural network architectures have been also utilized for this specific problem. CNNs are known for their strength in extracting local (e.g., spatial or temporal) features and inter-dependency of input nodes \citep{sainath2015deep}. In addition, the state-space model of the training variables is dramatically reduced since fixed sized kernels are being trained rather than large variable matrices from fully-connected layers. CNNs are mostly used for computer vision applications in which 2D kernels are applied on pixel pallets. In signal processing, 1D kernels are more proper choices. A dynamic response predictor for linear systems using CNNs is introduced in \citep{sun2017data}. More recently, \citet{wu2019deep} proposed a CNN-based algorithm for different partial dynamic response predictions. The most advanced case included prediction of acceleration response at the roof level of a multi degrees of freedom (MDOF) system given the ground motion. \par

Comprehensive dynamic response prediction of nonlinear systems has been investigated in a few recent studies. \citet{zhang2019physics} confirms that recurrent neural networks (RNN) are structurally great candidates for structural dynamic response modeling, however, technically they suffer from gradient-vanishing issue during training process. In fact, RNN models have been a frequently used architecture in the previously mentioned models (i.e., all one-step ahead response prediction models are basically RNN models). Based on this argument, \citet{zhang2019deep} proposes a long short-term memory (LSTM) architecture for the response modeling in order to address the gradient-vanishing issue. The primary difference of LSTM models compared to vanilla RNN models is the special architecture that allows for learning long-term temporal dependencies. This difference also handles the gradient-vanishing issue of RNN models. The study successfully predicted displacements, velocities, accelerations, and internal forces using the ground motion in different nonlinear cases. However, the model consisted of a large trainable variable space and required very long training process (i.e., 50,000 epochs). \par

The same research team has also recently introduced physics-guided models using CNN and LSTM architectures for dynamic response prediction problem \citep{zhang2019physics,zhang2020physics}. The studies propose an additional term in the loss function of the problem which penalizes deviations in the equation of motion when predicted outputs are plugged in. The studies showed that imposing this new physical constraint helped to enhance the prediction accuracy. Despite their high accuracy and completeness, the NN architectures are vanilla versions of the common NN types with no guidance from the physics. This results in over-complicated networks that require high number of training epochs. In addition, LSTM model requires a fixed signal length which is limiting. \par

In our study, we focus on designing architecture of a recurrent neural cell that updates the state from current time step to the next (i.e., one-step ahead predictor) with the neural connections that are inspired by exact numerical differential equation solvers. We believe that an ideal network is able to predict a response merely based on current time step of a full state space, as it is hardcoded in the simulation algorithms such as Newmark-$\beta$. 

\subsection{Motivation}

As the ubiquity of data-driven methods grows, the generalization and reliability of these models become more important. The vast majority of the available research train neural networks with no consideration for solid knowledge that governs the actual problem in hand. In addition, for engineering applications as opposed to data science problems, the available data is not extremely large and does not cover the entire domain of application possibilities (e.g., data is available for a limited domain of linear response in operational conditions). These two concerns demand for incorporating physics constraints into the architecture design of the NNs. On the other hand, as the problem holds more constraints, the training process eventually becomes harder. This study proposes a new approach to impose a special architecture that is inspired by implicit equation of motion solvers into a recurrent cell for full response prediction of nonlinear MDOF systems. The proposed network is called DynNet in this article, standing for dynamic network. Moreover, this study recommends multiple techniques so that the training process becomes smoother and more reliable. \par

\begin{figure}[!h]
    \centering
    \includegraphics[width=100mm]{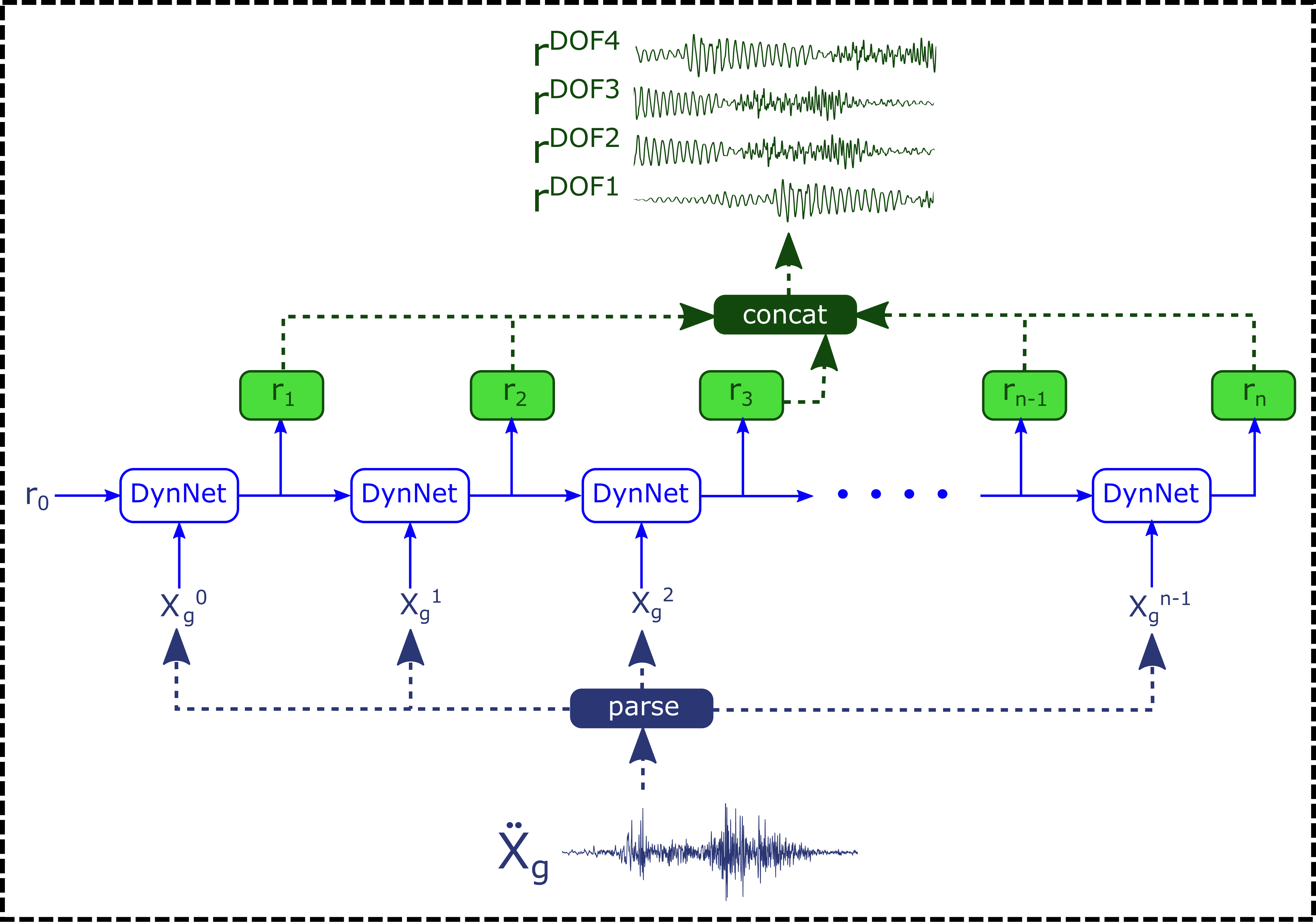}
    \caption{Schematic diagram of DynNet and conversion from ground motion to the structural response.}
    \label{fig:rnn_scheme}
\end{figure}

DynNet is a recurrent cell that performs one-step ahead prediction of the full state space of MDOF nonlinear dynamic system given a desired ground motion. The schematic structure of the network is presented in Figure \ref{fig:rnn_scheme}. This architecture has no limitation for the length of the signal. Our contribution is to design the architecture based on implicit dynamic simulation algorithms for nonlinear time history analysis (e.g., nonlinear Newmark-$\beta$ method). The key idea is that if the numerical algorithm is suitable and exact for nonlinear response analysis, a similar architecture has to be successful in learning the same nonlinear model from raw data. In addition, the architecture design is inspired by Residual Networks \citep{he2015deep} (i.e., ResNet) that have shown outstanding performances in learning partial differential equations from raw data. DynNet has significantly smaller dimension compared to the most accurate counterparts. \par

In terms of network optimization, this study utilizes second order trust region method which dramatically reduces required training iterations. Training dynamic blocks for one-step ahead prediction is highly sensitive to instability. To overcome this challenge, we introduce projection loss function. In addition, to accelerate learning ability of the network for nonlinear transitions, a hardsampling technique is proposed and implemented. Although DynNet is strongly constrained which results in harder training, its smaller variable space and high constraints enable network training with very limited amount of data. The physical interpretability of DynNet also helps to model highly severe nonlinear behaviors as well as very long signals, as we will show in the next sections. \par

In the following section, the detailed architecture of the network is elaborated. In Section 3 the technical approaches for faster and more robust training process of DynNet are presented (e.g., the optimization algorithm, loss function, and hardsampling technique). In Section 4 two numerical case studies are presented in which different types of nonlinearity are imposed. The summary of the method along with the highlights are presented in Section 5. 

\section{Physics-based Neural Network Architecture Design}\label{sec:nn}

\subsection{Numerical Solution for Direct Problems}

For simulation of dynamic systems, implicit solvers analyze responses at time step $i$ to derive response at time step $i+1$. In fact, regardless of the complexity and level of nonlinearity of the problem, simulators require no further information for one-step ahead prediction. Relying on this fact, an ultimate simulator that learns from data should be a dynamic cell that is able to perform one-step ahead prediction with high accuracy and low cumulative error. In addition, considering the causality of the dynamic system as well as its short memory (i.e., a few recent samples are sufficient for the next step prediction), LSTM models seem unnecessarily over-complicated. DynNet is a robust one-step ahead dynamic cell that is very sharp in learning nonlinearities as well as robust to noise. In this study, we do not use a simplified version of existing networks such as CNN or LSTM, but instead design the internal cell connections in a way that conforms with common dynamic simulation solvers. The nonlinear version of Newmark's algorithm is shown in Algorithm \ref{alg:newmark} \citep{riddell1979statistical}.\par

\begin{algorithm}[!ht]
	\caption{Newmark's Method for Nonlinear Systems.}
	\label{alg:newmark}
	\small \small
	\begin{algorithmic}[1]
		\State \textbf{Input}: $u_i,\dot{u_i},\ddot{u_i},S_i,\ddot{x}_i^g$, TangentStiffness$(.)$, NonlinearForce$(.)$
		\State $a_1,a_2,a_3,C_1,C_2,C_3,C_4,C_5,C_6,M,\Gamma:=$ Constant
		\State $\hat{p}_{i+1} = M \Gamma \ddot{x}_i^g + a_1 u_i + a_2 \dot{u_i} + a_3 \ddot{u_i}$
		\State $R^{(0)} = \hat{p}_{i+1}$
		\State $j = 0$
		\State $K^t_i = $TangentStiffness$(u_i,\dot{u_i},\ddot{u_i},S_i)$
		\While {$abs(R^{(j)}) < threshold$} 
		\State $R^{(j)}= \hat{p}_{i+1} - S_{i+1}^{(j)} - a_1 u_{i+1}^{(j)}$
        \State $(K^t_{i+1})^{(j)} = (K^t_{i+1})^{(j)} + a_1$
        \State $\Delta u^{(j)} = ((K^t_{i+1})^{(j)})^{-1} R^{(j)}$
        \State $u_{i+1}^{(j+1)} = u_{i+1}^{(j)} + \Delta u^{(j)}$
        \State $S_{i+1}^{(j+1)} = $ NonlinearForce($u_{i+1}^{(j+1)},S_i^{(j)}$)
        \State $j = j+1$
        \EndWhile
        \State $u_{i+1} = u_{i+1}^{(j)}$
        \State $S_{i+1} = S_{i+1}^{(j)}$
        \State $\dot{u}_{i+1} = C_1 (u_{i+1} - u_i) + C_2 \dot{u_i} + C_3 \ddot{u_i}$
        \State $\ddot{u}_{i+1} = C_4 (u_{i+1} - u_i) + C_5 \dot{u_i} + C_6 \ddot{u_i}$
        \State \textbf{Return} $u_{i+1},\dot{u}_{i+1},\ddot{u}_{i+1},S_{i+1}$
	\end{algorithmic}
\end{algorithm}	

In this algorithm, $u_i,\dot{u_i},\ddot{u_i}$ are displacement, velocity, and acceleration vectors of current time step $i$, respectively. $S_i$ and $\ddot{x}_i^g$ are respectively the internal force vector and ground motion acceleration at time $i$. In this algorithm, the detailed expressions for constant coefficients are discounted. The algorithm consists of a majority of linear expressions and some nonlinear functions - TangentStiffness(.) and NonlinearForce(.) - that depend on the defined nonlinearity of the system (the first function returns tangent stiffness and the second function derives nonlinear story forces based on the nonlinear model). In particular, the algorithm can be divided into three blocks: (a) initialization; (b) equilibrium solver; and (c) post-processing. In this organization, blocks (a) and (c) merely include linear operations. For instance, in Line 16, the relationship between displacement and velocity of the future time step is a linear expression.\par

In addition, block (b) contains a \texttt{while} loop which certifies the equilibrium (i.e., Newton-Raphson root finding solution). Intuitively, this while loop incrementally adds up values to its estimation of $u_{i+1}$ every time the loop runs. This mechanics resemble the mechanics of Residual Networks (ResNet) \citep{he2015deep} in which the output of the network is added to the input and fed back to the network repeatedly. Studies have shown that ResNets outperform other architectures in learning differential equations from data \citep{lu2017beyond,chen2018neural} due to their inherited resemblance to the Euler's method. \par

\begin{figure}[!h]
    \centering
    \includegraphics[width=145mm]{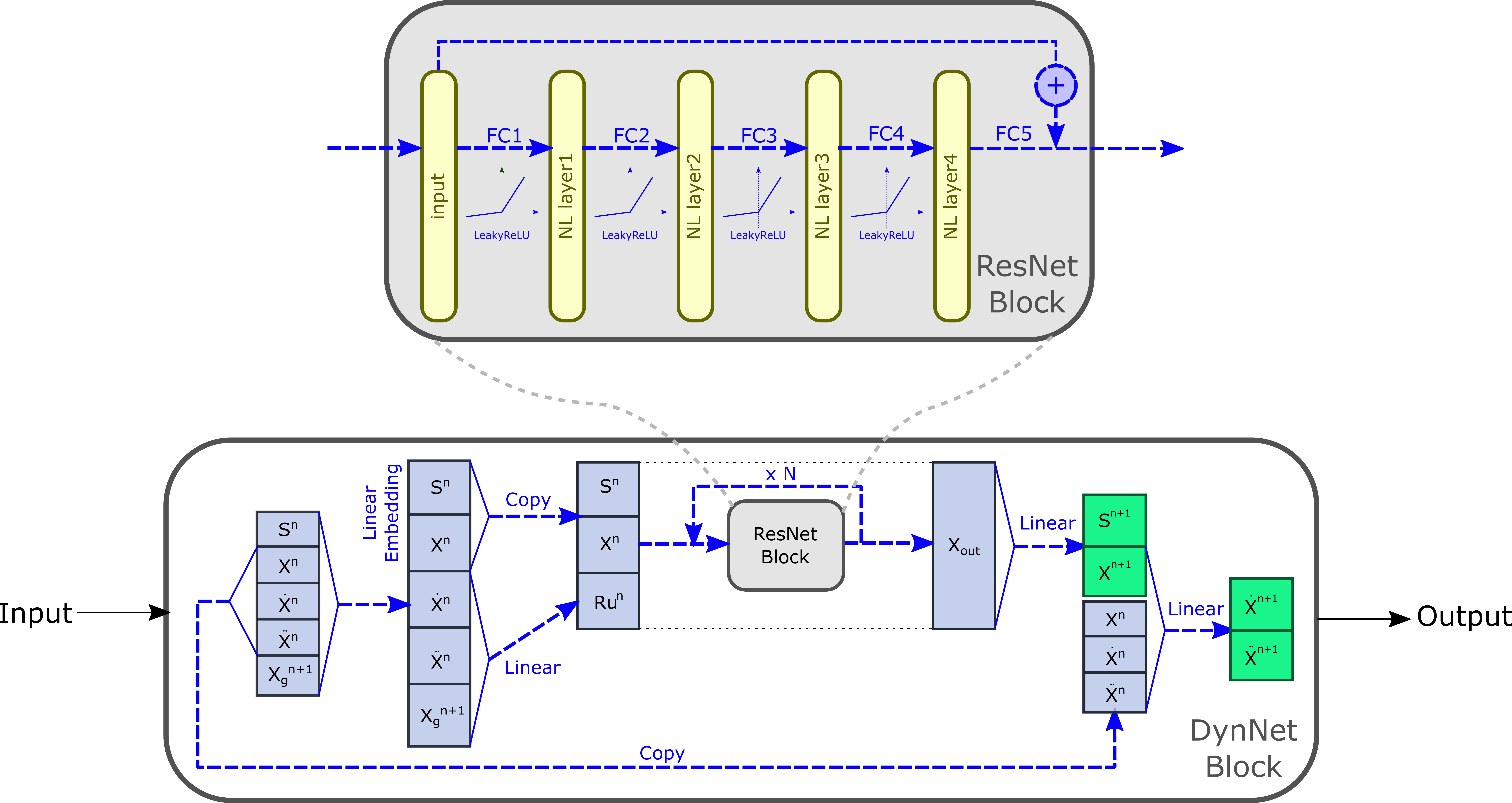}
    \caption{DynNet recurrent cell components.}
    \label{fig:dynnet_blocks}
\end{figure}

\subsection{DynNet Components}

DynNet is designed to benefit from two intuitive ideas: (1) inspired by the structure of numerical implicit simulators; and (2) ResNet structure for nonlinearity learning. The architecture of the network is given in Figure \ref{fig:dynnet_blocks}. The input of the network is identical to the Newmark's algorithm. All connections in the network are linear expect for the internal connections of the ResNet block. The network initially adjusts the dimension of the input vector via a linear embedding layer. Then, velocity and acceleration of the structure in addition to the ground motion acceleration of the current time step are fed into a linear layer to produce $R_u^n$ (equivalent to $R$ in Algorithm \ref{alg:newmark}). Then, internal force, displacement, and $R_u$ are concatenated and passed into the ResNet block. The ResNet block is expanded in Figure \ref{fig:dynnet_blocks} as well. This block is the sole component of the network that is able to learn the nonlinear behavior of the dynamic system. The block is conveniently arranged with stacked fully-connected layers that are connected with leaky rectified linear units (i.e., LeakyReLU activation functions). The output of the fifth fully-connected layer is added to the input of the ResNet block to produce the terminal state of the ResNet block. This terminal state is fed back to the ResNet block $N$ times ($N$ is a user defined parameter). After $N$ repetitions, the output is linearly mapped to $S_{i+1}$ and $X_{i+1}$. Given $X_{i+1}$, the velocity and acceleration of the next time step are derived by another linear map. Once the prediction of time step $i+1$ is found, it will be fed back to DynNet for the response prediction of the consecutive time step (e.g., $i+2$).\par

The concentrated learning ability that is placed in the ResNet block enables easy replacement of the simple MLP network with other nonlinear structures (e.g., CNN or deeper networks). This feature decouples the nonlinearity learning and state transitioning tasks of the network. In other words, for very involved types of nonlinearities, one simply requires to modify the structure of the ResNet block (e.g., add extra layers). However, in this study we found a five layer MLP sufficiently strong for the test cases. The variable space of the network is highly dependent to the user-defined embedding dimension. In this study, embedding size is set to eight for all cases, yielding $5,320$ trainable variables. The dimension is significantly lower compared to other recently developed networks for the same purpose. \par

\section{Accelerating Techniques for the Training Phase}

\subsection{Selecting Optimizer}

Stochastic first-order methods, including SGD  \citep{robbins1951stochastic} and Adam \citep{kingma2014adam}, are currently standard optimization methods for training neural network problems. These methods have a low per-iteration cost, enjoy optimal complexity, and are easy to implement and applicable to many machine learning tasks. However, these methods have several issues: (i) they are highly sensitive to the choice of hyper-parameters (such as batch size and learning rate); and more importantly (ii) they are not effective for ill-conditioned problems, meaning that for a small change in the inputs, the outputs can change dramatically. The second issue is quite likely when dealing with nonlinear structural systems. For instance, in an elasto-plastic model, there is a bounded relationship between force and displacement within the elastic range. However, the variations of displacements become extremely large when the system experiences larger forces (i.e., forces beyond the elastic limit).

\begin{figure}[!h]
    \centering
    \includegraphics[width=120mm]{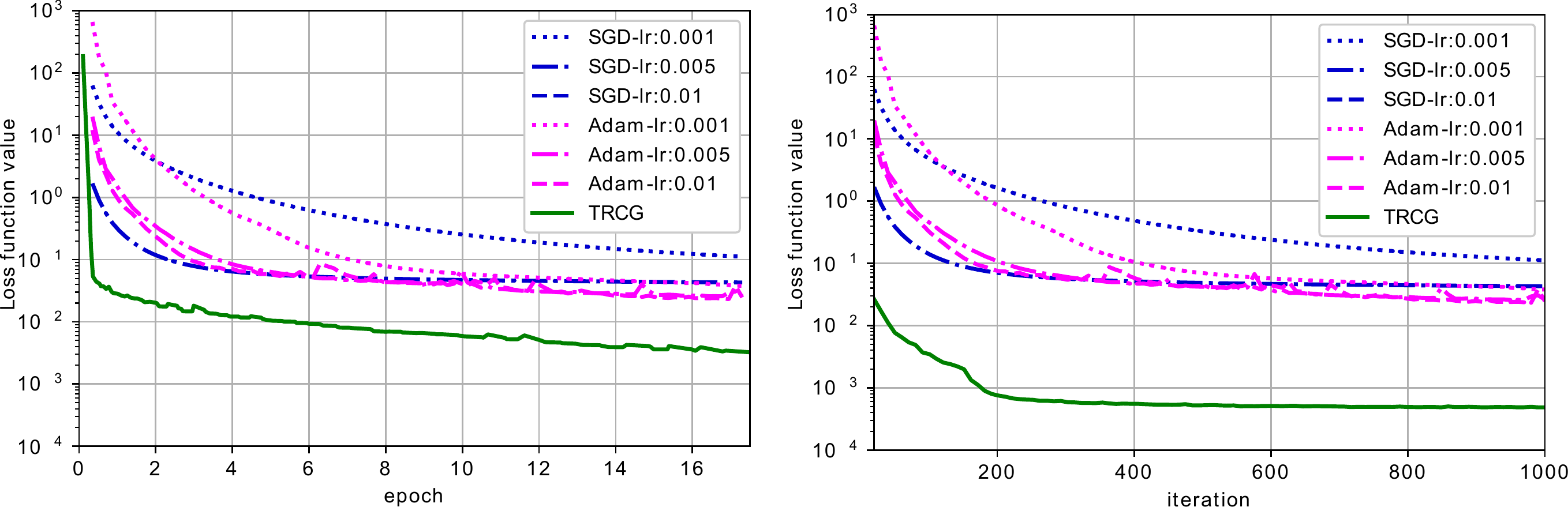}
    \caption{Optimization trends using different optimizers.}
    \label{fig:optimizers}
\end{figure}

On the other hand, second-order methods by utilizing second-order (i.e., curvature) information can address the aforementioned issues. One class of second-order methods are Hessian-free methods, in which no Hessian is needed to be constructed explicitly, and only Hessian-vector multiplications are needed in order to update the neural network parameters. In our study, we utilize a method in the Hessian-free class which is called \textit{Newton trust-region approach} (TRCG). This is motivated by the results presented in Figure \ref{fig:optimizers} that illustrate the performance of TRCG and some of the well-known stochastic first-order methods with different choices of hyperparameters. As is clear from the results, the performance of TRCG by utilizing the curvature information is noticeably better than the stochastic first-order methods in terms of loss function value with respect to both iteration and epoch number. Similar behaviour is also observed in \citep{berahas2019quasi,xu2020second}. In every iteration of TRCG, the following non-convex quadratic sub-problem needs to be solved:  

\begin{equation}
    \begin{aligned}
       p_k \in 
       \argminA_{p \in \mathbb{R}^d} \; Q_k(p) = &p^T g_k + \tfrac{1}{2}p^T H_k p \\
       \text{s.t. } & \|p\|  \le \Delta_k,
    \end{aligned}{}
\end{equation}
where $g_k$ is the (stochastic) gradient, $H_k$ is the (stochastic) Hessian, and $\Delta_k$ is the trust-region radius at iteration $k$.
The above sub-problem can be approximately and efficiently solved using CG-Steihaug \citep{nocedal_book} which is summarized in Algorithm \ref{alg:CG-Steihaug}. The output of Algorithm \ref{alg:CG-Steihaug}, $p_k$, is the search direction in order to update the neural network parameters. In other words, assume we are at $k^{\text{th}}$ iteration, and the neural network parameters are updated as $\omega_{k+1} := \omega_{k} + p_k$. More details regarding the trust-region algorithm, the strategy for updating $\Delta_k$, accepting or rejecting the steps can be found in \citep{nocedal_book}.
\begin{algorithm}[!h]
\small 
\caption{CG-Steihaug \citep{nocedal_book}.}
  \label{alg:CG-Steihaug}
 {\bf Input:} $\epsilon$ (termination tolerance), $g_k$ (current gradient).

  \begin{algorithmic}[1]
    \State Set $z_0 =0$, $r_0= g_k$, $d_0 = -r_0$
   \If{ $\| r_0\| < \epsilon$}
    \State {\bf return} $p_k = z_0 = 0$
  \EndIf
  \For {$j=0,1,2,...$}
     \If{ $d_j^TH_kd_j \leq 0 $}
        \State Find $\tau \geq 0$ such that $p_k = z_j + \tau d_j$ minimizes $m_k(p_k)$ and satisfies $\|p_k\| = \Delta_k$
        \State {\bf return} $p_k $
    \EndIf
    \State Set $\alpha_j = \dfrac{r_j^Tr_j}{d_j^TH_kd_j}$ and $z_{j+1}= z_j + \alpha_jd_j$
     \If{ $\| z_{j+1}\| \geq \Delta_k$}
        \State Find $\tau\geq 0$ such that $p_k = z_j + \tau d_j$ and satisfies $\|p_k\| = \Delta_k$
        \State {\bf return} $p_k $
    \EndIf
    \State Set $r_{j+1}= r_j + \alpha_jH_kd_j$
    \If{ $\| r_{j+1}\| < \epsilon_k$}
        \State {\bf return} $p_k = z_{j+1}$
    \EndIf 
    \State Set $\beta_{j+1} =\dfrac{r_{j+1}^Tr_{j+1}}{ r_{j}^Tr_{j}} $ and $d_{j+1} =-r_{j+1} + \beta_{j+1}d_j $
    \EndFor
  \end{algorithmic}
\end{algorithm}

\subsection{Projection Loss}

In order to train a recurrent block for one-step ahead prediction, the simplest approach is to minimize the residue between the predictions and the actual values over a mini-batch in each iteration. However, this approach for training produces very unstable networks, which are prone to divergence when predicting a long trajectory of responses given the initial conditions. To address this issue, we introduce and utilize projection loss that is the basis for the training process in this study. \par

\begin{figure}[!h]
    \centering
    \includegraphics[width=60mm]{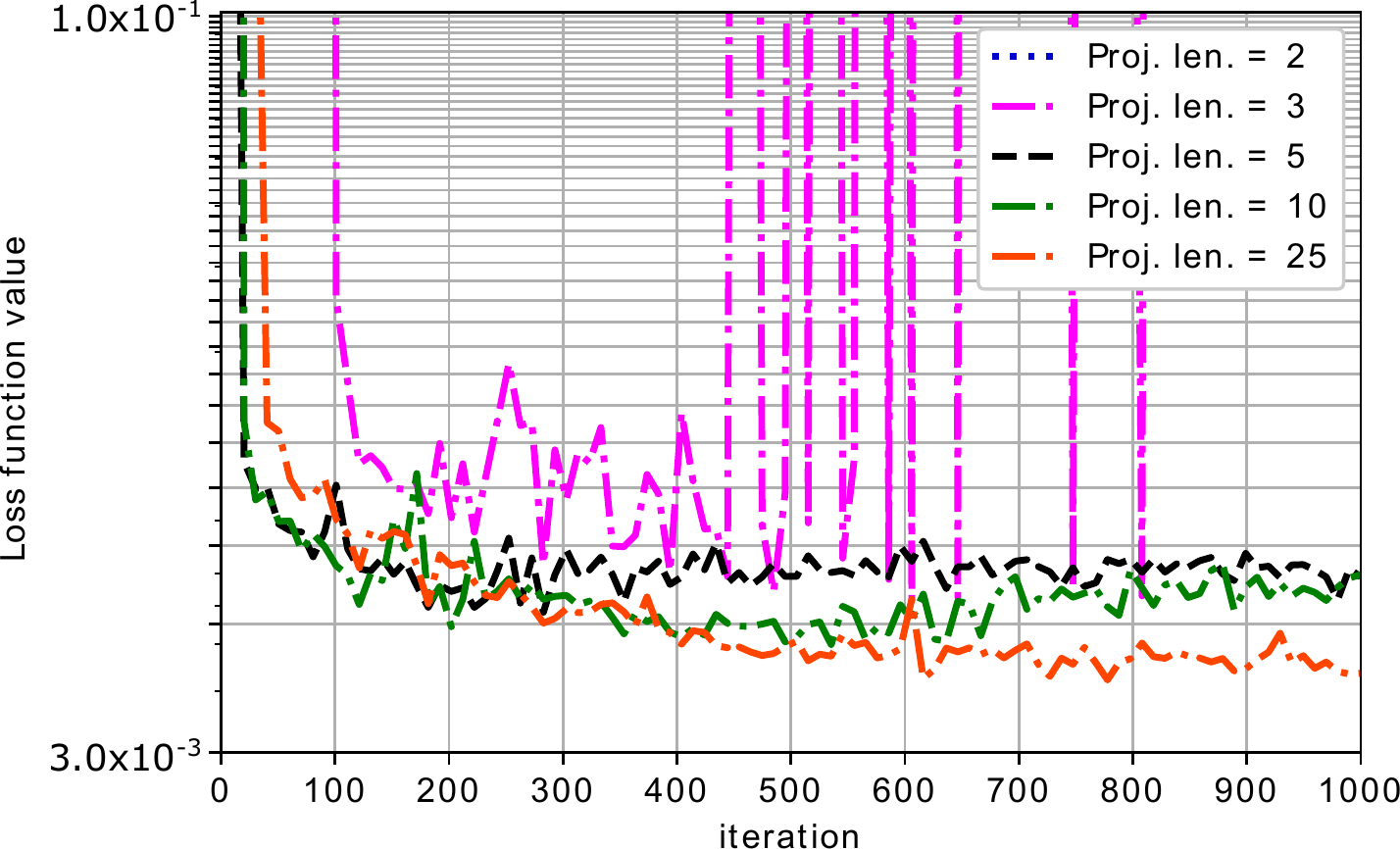}
    \caption{Optimization trends using different projection lengths.}
    \label{fig:projlen}
\end{figure}

Projection loss is calculated as the mean squared error of a sequence of responses predicted by DynNet when compared with the corresponding actual responses. To produce the sequence of predicted responses, the only given value is the initial conditions at some randomly selected time step. This initial condition is then fed into the DynNet and the responses are fed back for $N$ times to predict a trajectory starting from the random initial condition ($N$ is a user-defined projection length). Compared to the conventional loss function, the projection loss can effectively control the instability issue of the neural network. Figure \ref{fig:projlen} demonstrates the effect of loss functions with different projection lengths on the testing loss. \par

As shown in Figure \ref{fig:projlen}, the length of the projection directly affects the robustness of the optimization. In fact, when the projection length is two, the network's inference diverges (i.e., after multiple steps of recurrence, DynNet outputs explode and it is outside the shown range in the figure). The best results on the testing data are observed when the projection length equals to 25. Note that as the projection length in the loss function increases, the model becomes more optimal for longer trajectory predictions, however, the training time linearly increases as well. In fact, for loss functions with longer projection lengths, the forward pass and backpropagation steps take longer and these computations cannot be distributed over the processing resources (due to the sequential nature of the network inference). In addition, by comparing results from projection length = 5 and projection length = 25, it is observed that the former performs better initially (i.e., in lower iterations) while the latter shows its advantage later on. From this observation, we adopt a sequentially increasing projection length model in this study. In the following section, the models are trained for loss functions with projection lengths equal to 5, 10, and 25, respectively; for each, the models are trained for a fixed number of iterations. 

\subsection{Hardsampling Technique}

For learning highly nonlinear systems, samples may be distributed extremely unevenly in different behavioral regions. For instance, elasto-plastic systems normally respond linearly to the major portion of a ground motion, regardless of the intensity of the motion. In other words, the system undergoes nonlinear deformations occasionally when a large impact occurs in the input. As a result, the portion of one-step ahead response transitions that are within elastic region is dramatically larger than the inelastic region. This induces a severe imbalance in the training data distribution, which turns out to be detrimental for model's robustness. Importance sampling is a technique for online batch selection that is used to circumvent the problem with unevenly distributed data. \par

A review of more common batch selection methods are given in Section 7 of  \citep{loshchilov2015online}. One of the simplest and most effective approaches for adaptive batch selection is rank-based selection \citep{schaul2015prioritized,loshchilov2015online}. In this method, during the training phase, samples of each batch are sorted in descending order based on their function value, and then, their probability of re-selection is updated based on their ranking. The idea was first employed for reinforcement learning using temporal difference (TD) as the reference for sample sorting, and later was adopted for deep learning applications and based on loss function value. In this study, a similar approach is introduced which is inspired by the notion of ranked-based batch selection. \par

In the implemented hardsampling technique, a hardsampling rate $r$ is defined which is the proportion of samples in the batch that are eventually selected from the hardsamples. The model starts with randomly selected samples in the first iteration. At the end of the iteration, the $N$ (is a user-defined hyperparameter) samples with the maximum contributions in the total batch loss are added to a list of hardsamples. In fact, the list of hardsamples is a bag of samples that are not learned well by the model yet. In the next iteration, the batch samples are selected such that $b_1$ samples are randomly selected from the entire training samples and $b_2$ samples are randomly selected from the the list of hardsamples and $b2 = \floor{r\times (b_1+b_2)}$. At the end of the iteration, the list of hardsamples is updated and passed to the next iteration. The process continues accordingly. \par

\begin{figure}[!h]
    \centering
    \includegraphics[width=60mm]{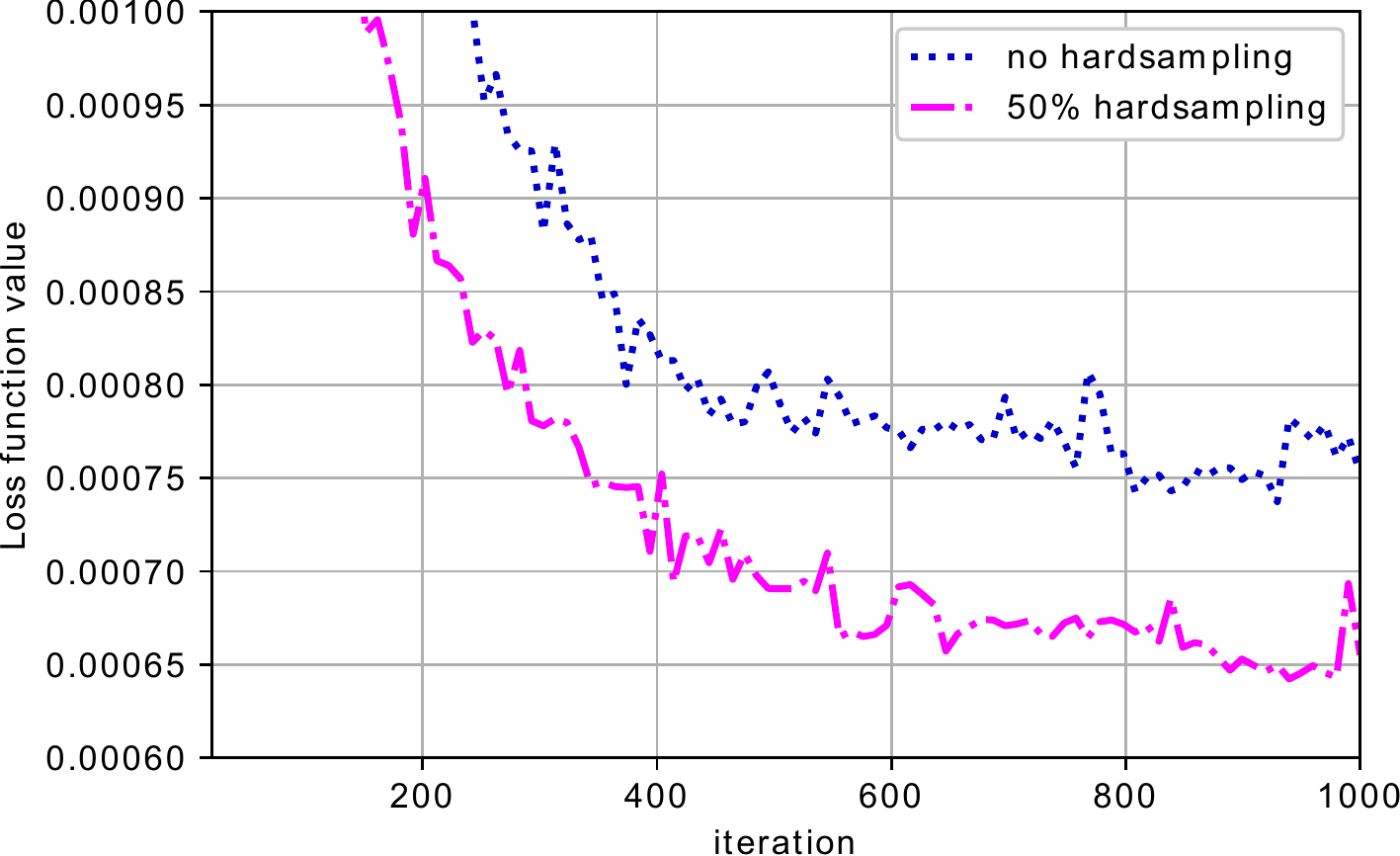}
    \caption{Optimization trends using different hard sampling ratios.}
    \label{fig:hardsampling}
\end{figure}

To evaluate the effectiveness of the technique, the optimization process is performed with and without hardsampling technique and results are compared in Figure \ref{fig:hardsampling}. In this example, the rate of hardsampling $r$ is $50\%$. The result clearly confirms the advantage of hardsampling technique in fast learning and better learning of the model. Therefore, in this study this technique is also used in the training process of the models. The approach is adaptive, meaning that the training process automatically picks hardsamples throughout the training process. In engineering problems, we may have an \textit{a-priori} hypothesis about the hardsamples. For instance, in the elastoplastic models, it is expected that one-step ahead response transitions that are beyond the elastic limit are hardsamples. In the next section, we will confirm that our adaptive hardsampling technique automatically detects these samples. 

\section{Numerical Case Studies} \label{sec:casestudy}

\begin{figure}[!h]
    \centering
    \includegraphics[width=60mm]{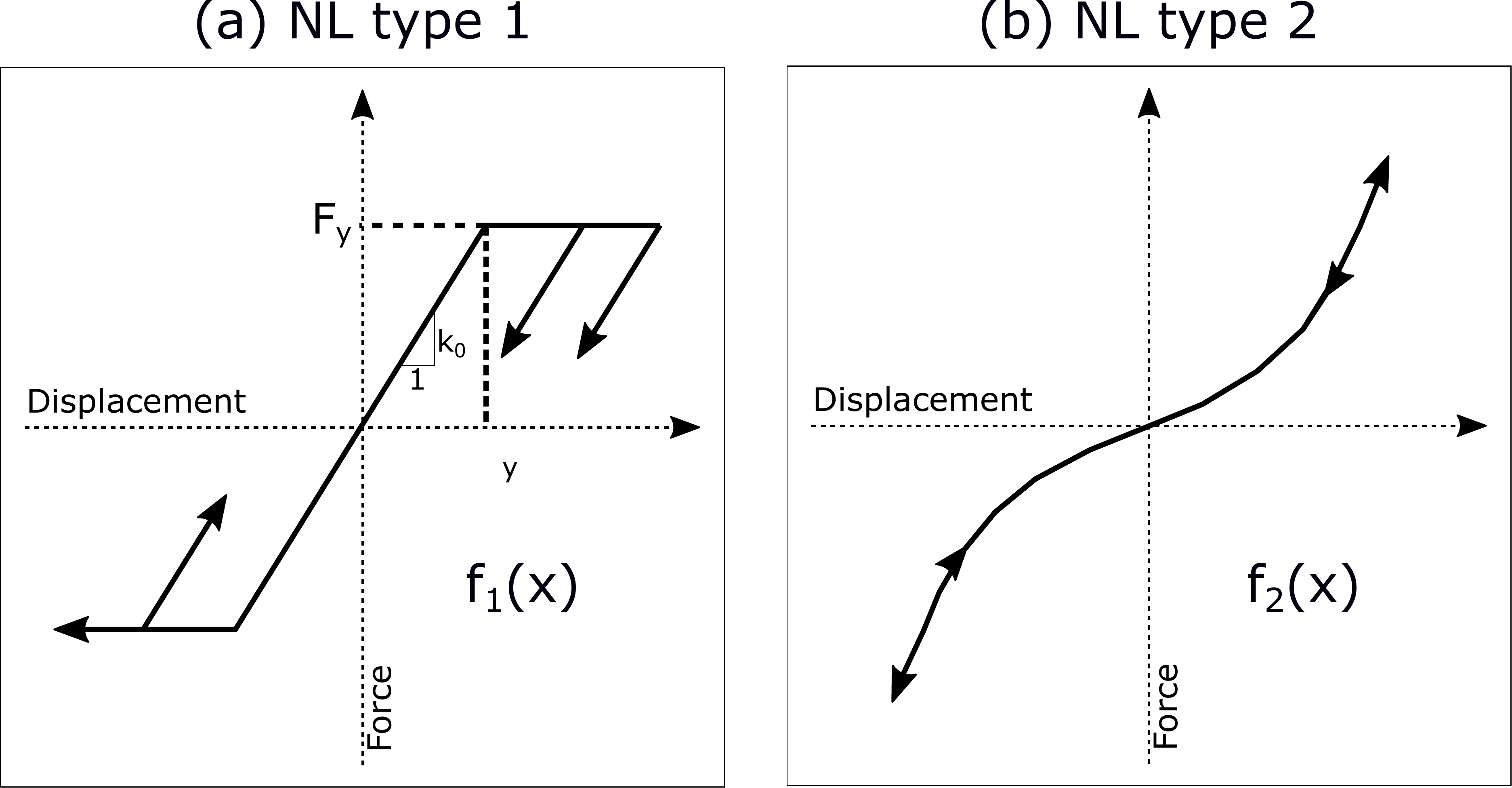}
    \caption{Force-displacement relationships of two nonlinear cases.}
    \label{fig:forcedisp}
\end{figure}

In this section, two case studies are considered to validate the strengths of DynNet in response prediction of different nonlinear systems. These case studies differ in terms of the type of introduced nonlinearity to the systems. The first case is a four degrees of freedom (DOF) system with elastic perfectly plastic springs. The second model consists of a four-DOF system equipped with nonlinear ($3^{rd}$ order) elastic stiffeners (schematics of the force displacement behaviors are  shown in Figure \ref{fig:forcedisp}). The governing equations of motion (EOM) for these two nonlinear systems are shown in Equations \ref{eq:eom} and \ref{eq:functions}. 


\begin{equation}
    \begin{aligned}
       m\ddot{x} + c\dot{x} + f(x) = -m \Gamma \ddot{x_g}. \\
    \end{aligned}{}
    \label{eq:eom}
\end{equation}
        
\begin{equation}
    \begin{aligned}
        f_1(x) = 
         \begin{cases} 
         k_0x & x\leq \Delta_y, \\
         F_y & x > \Delta_y. \\
        \end{cases} \\
       f_2(x) = k_1 x + k_2 x^3.
    \end{aligned}{}
    \label{eq:functions}
\end{equation}

\begin{figure}[!h]
    \centering
    \includegraphics[width=60mm]{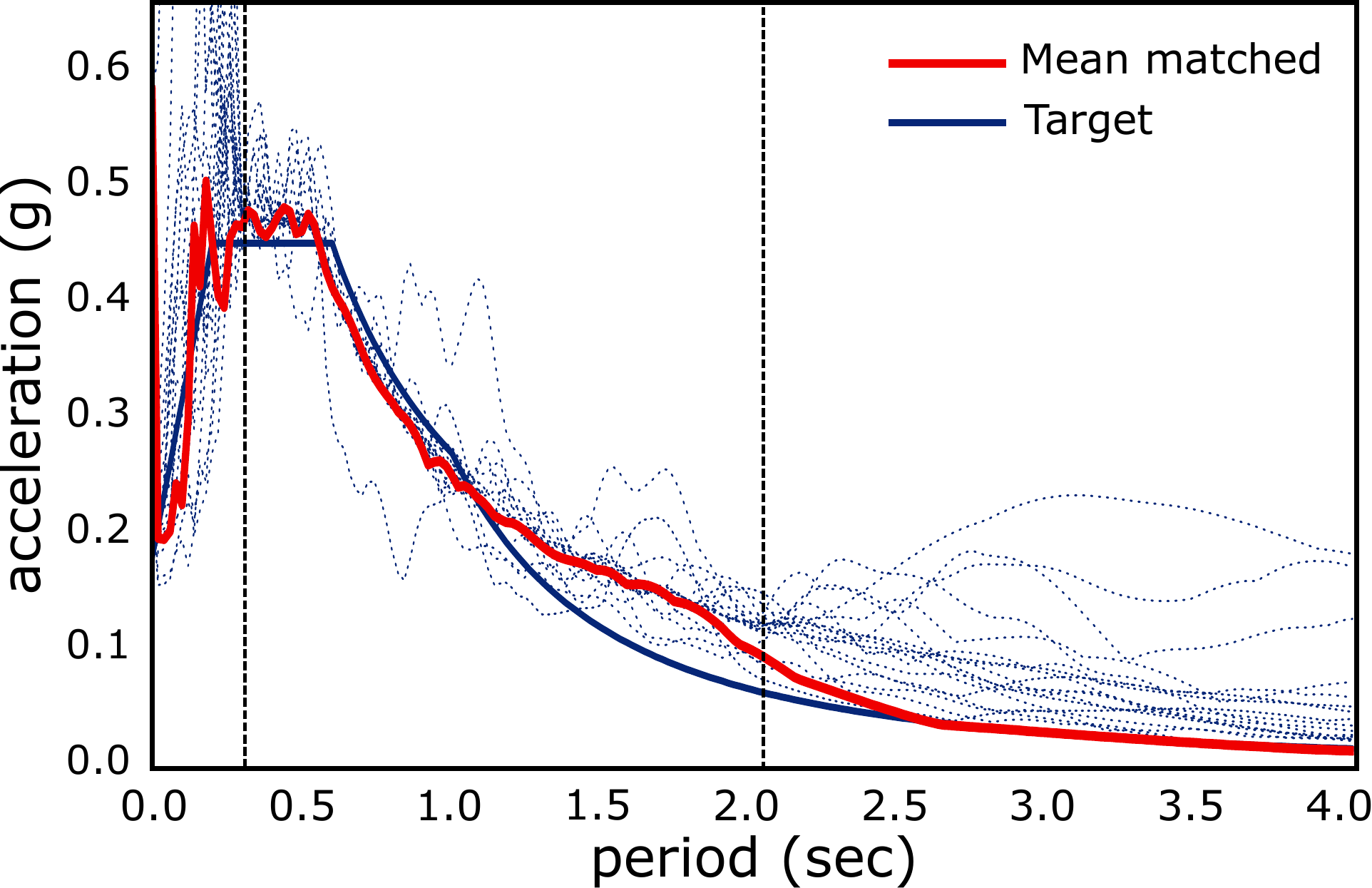}
    \caption{Earthquake response spectra matched with respect to the target spectrum and the mean spectrum.}
    \label{fig:GM_scaling}
\end{figure}

For the numerical simulation, Newmark's method for nonlinear systems is used in MATLAB. For this purpose, 20 strong ground motions are randomly selected from Center for Engineering Strong Motion Database (CESMD) \citep{haddadi2008center}. In addition to that, 10 band limited random time series are synthesized and added to the the library of input signals. The earthquake ground motions are scaled using the wavelet algorithm proposed by \citet{hancock2006improved}. The target matched spectra for twenty earthquake ground motions as well as the mean matched and target spectra are shown in Figure \ref{fig:GM_scaling}. The algorithm scales the time histories in a way that the response spectrum optimally matches with the target spectrum within the range of $0.2T_1$ to $1.5T_1$ ($T_1$ is the structure's natural period of the first mode).

For each case study, the scaled earthquake ground motions as well as random time histories are simulated to predict structure's responses (i.e., displacement, velocity, and acceleration) at all four DOFs. This data include both training and testing datasets. From 30 simulated ground motions, eight ground motions are randomly picked to be used as the training dataset and the rest for testing. Note that since DynNet is heavily constrained by the physics of the problem and enjoys low training variable space, it is expected that the model is easily trainable with small amount of training data and also is desirably generalized for a wide range of testing data. \par

\subsection{Case 1: Elastic-Perfectly Plastic Model (NL type 1)}\label{ssec:case1}

In this section, the results on the first test case - a four DOF shear building with elastic-perfectly plastic stiffness - are presented. The mechanical properties of the structure is presented in Table \ref{tbl:nltype1}. In this table, $M1-M4$ and $K1-K4$ stand for mass and elastic stiffness values of DOF1 to DOF4, respectively. $Fy$ shows the stories' yielding force. To consider the robustness of DynNet, three levels of noise are also considered ($0\%$, $5\%$, and $10\%$ noise levels). The network is trained to predict the full response at all DOFs including displacement, velocity, and acceleration time histories given the earthquake ground motion. \par

\begin{table}[!h]
    \centering
    \caption{Mechanical properties for NL type 1.}
    \begin{tabular}{l l l}
    \toprule
    \textit{Mechanical props.} & \textit{Values} & \textit{Units}\\
    \midrule
    M1                                                                & 0.259  & $kip.s^2/in$ \\ \hdashline  
    M2/M1                                                             & 1      & -                           \\\hdashline  
    M3/M1                                                             & 0.75   & -                           \\\hdashline  
    M4/M1                                                             & 0.5    & -                           \\\hdashline  
    Fy                                                                & 50     & $kips$                        \\\hdashline  
    K1                                                                & 168    & $kips/in$                     \\\hdashline  
    K2/K1                                                             & 7/9    & -                           \\\hdashline  
    K3/K1                                                             & 1/3    & -                           \\\hdashline  
    K4/K1                                                             & 1/4    & - \\                         
    \bottomrule
    \end{tabular}
    \label{tbl:nltype1}
\end{table}

\begin{figure}[!h]
    \centering
    \includegraphics[width=120mm]{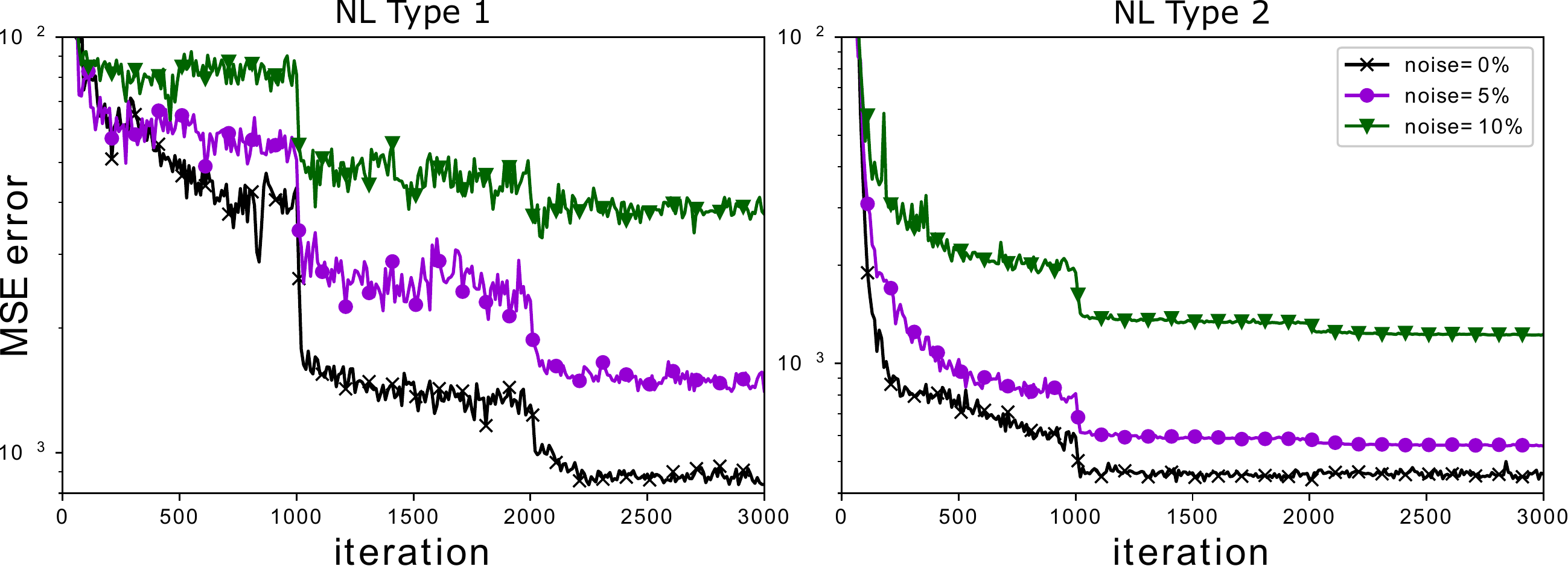}
    \caption{Loss function reduction versus iteration: the projection length for loss function calculation changes at iteration 1000 and 2000 (length equals to 10, 25, and 50 for each portion). The sudden drops in the loss function values at those iterations show the effectiveness of the proposed training technique.}
    \label{fig:lossTrends}  
\end{figure}

As concluded in the previous section, the network is trained in a multilevel manner: 1000 iterations with 10-step projection loss, then 1000 iterations with 25-step projection loss, and finally, 1000 iterations with 50-step projection loss. During the training process, batch size was set fixed at 1024 (i.e., 1024 one-step ahead transitions). In total, the network is trained for less than 100 epochs using TRCG optimizer. The learning curve is presented in Figure \ref{fig:lossTrends} (nonlinear (NL) type 1). The figure demonstrates that by increasing the length of projection in the custom loss function, a sharp drop in the loss function occurs. \par

\begin{figure}[!h]
    \centering
    \includegraphics[width=120mm]{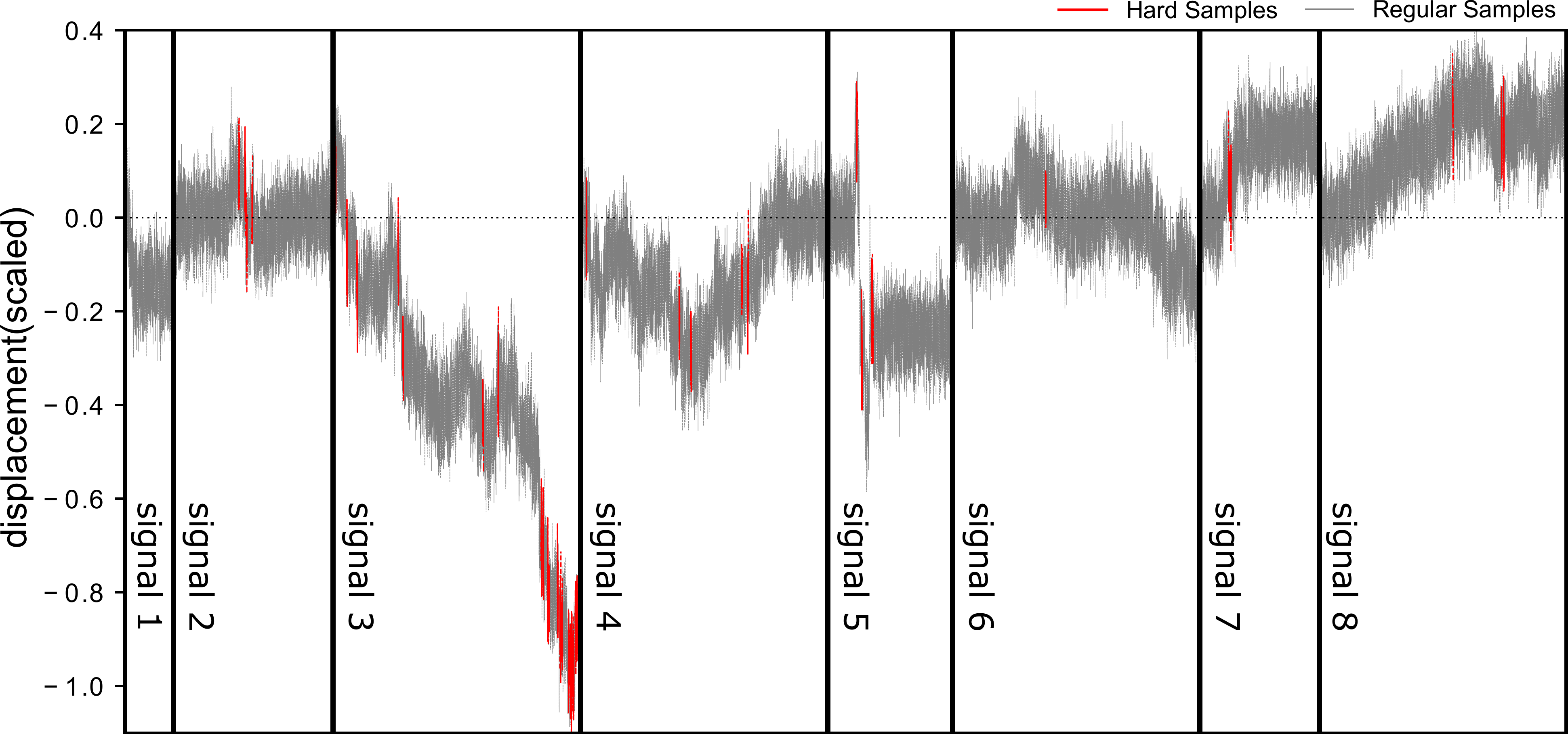}
    \caption{Locations of hard samples for adaptive sampling in the nonlinearity type 1: as expected, the majority of hard samples are located when large residual displacements occur.}
    \label{fig:hardsamples}
\end{figure}

As previously explained, the training phase incorporates the proposed hardsampling technique. To evaluate the physical interpretation of automatically selected hard samples, Figure \ref{fig:hardsamples} is presented. In this figure, the entire training dataset (including eight signals) are shown and divided by vertical lines. The signal portions that are labelled as hardsamples are color coded in red. Interestingly, hardsamples are mostly found when a sudden drop (due to a severe nonlinear behavior) has happened. This observation confirms that the algorithm reuses highly nonlinear samples to intensify its learning ability. \par

\begin{figure}[!h]
    \centering
    \includegraphics[width=145mm]{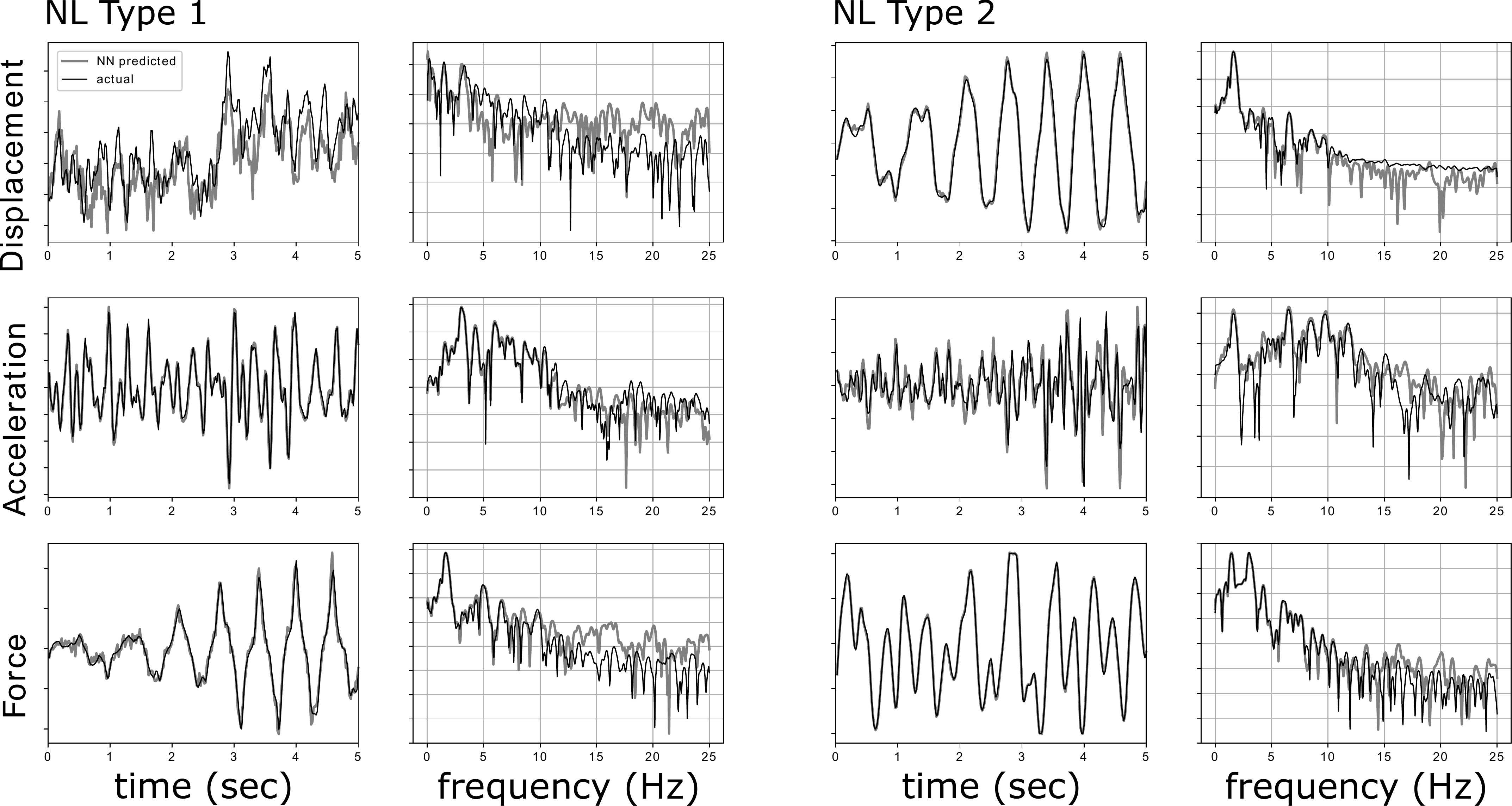}
    \caption{Predicted signals for 5 seconds with 5\% noise. The plots show that the network is very accurate in predicting responses for a short future. The same level of accuracy is visible in both time and frequency representations of the signals.}
    \label{fig:sig250}
\end{figure}

\begin{figure}[!h]
    \centering
    \includegraphics[width=145mm]{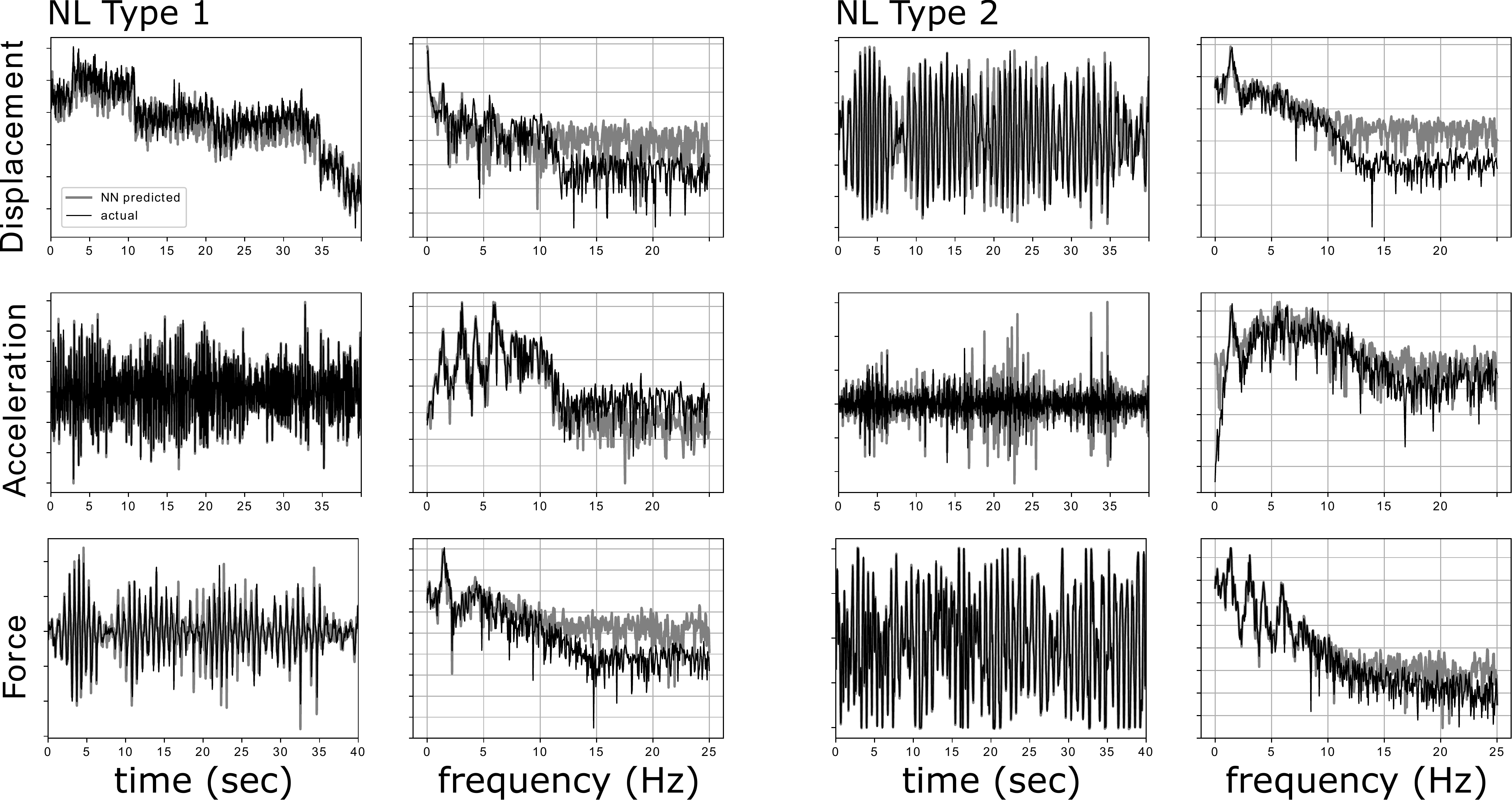}
    \caption{Predicted signals for 40 seconds with 5\% noise. The plots show that the network is still accurate in predicting responses for a longer time. The same level of accuracy is visible in both time and frequency representations of the signals. Notice that the displacement prediction for the NL type 1 is strongly nonlinear. However, the network successfully estimated it.}
    \label{fig:sig2000}
\end{figure}

To evaluate the prediction performance of the trained network, the prediction results on one randomly picked testing signal with $5\%$ noise for short and long trajectories are presented in Figures \ref{fig:sig250} and \ref{fig:sig2000}. The predictions are compared with the reference signals in both time and frequency domains (velocity predictions are neglected for brevity). For short trajectories (i.e., five second prediction in Figure \ref{fig:sig250}), the performance is promising. Note that the nonlinear baseline variations are accurately predicted in the displacement time signals. In terms of frequency, the accuracy of predicted signal is very high. For longer trajectories (i.e., 40 second prediction in Figure \ref{fig:sig2000}), the prediction accuracy is as high. The modal peaks in frequency domain are captured accurately. Notably, all the baseline variations in the displacement time signal are predicted accurately using the trained network. Such high accuracy for predicting severely nonlinear responses are unprecedented in the literature. \par

\begin{figure}[!h]
    \centering
    \includegraphics[width=120mm]{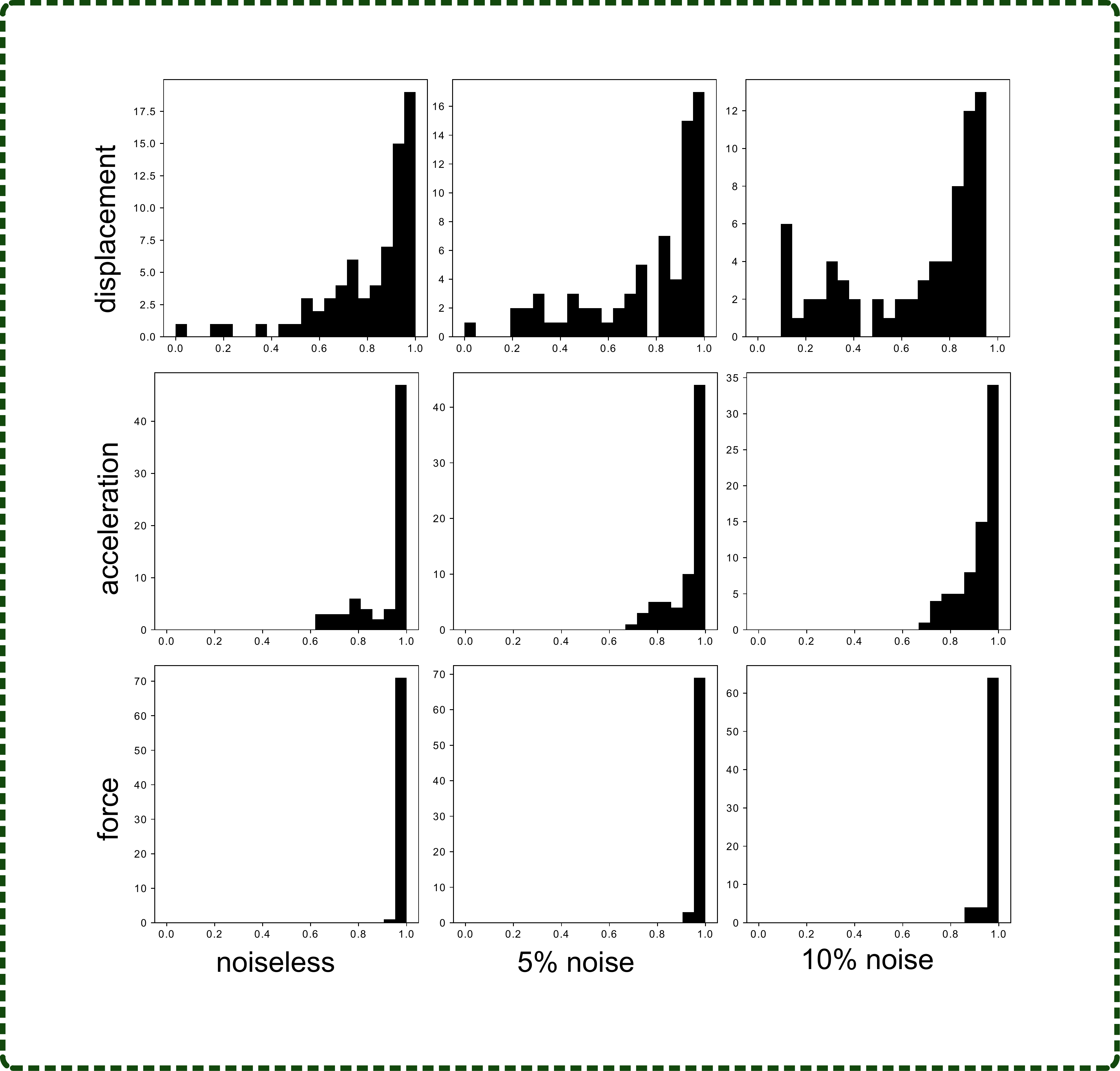}
    \caption{Pearson correlation coefficient histogram for predicted responses - nonlinearity type 1.}
    \label{fig:hist_NL1}
\end{figure}

To further quantify the accuracy of the predictions in all the testing signals, Pearson correlation coefficients (PCC) are calculated between predicted and ground truth signals (40 second predictions) and presented in Figure \ref{fig:hist_NL1}. PCC is a measure to quantify the fitness of predicted trajectories with respect to the ground truth signals \citep{weisstein2006correlation}. The results for all three noise levels are presented. The histograms demonstrate the distribution of different prediction accuracy. In general, for all predicted quantities (i.e., displacement, acceleration, and internal force) and all noise levels, more than $90\%$ of DynNet's predictions have PCC above $0.8$. Particularly, force and acceleration predictions are exceptionally accurate. Note that in the noisy cases, the likelihood of having very high PCC is inevitably low due to the irreducible noise. Still, DynNet shows a very good performance in response predictions subjected to these highly nonlinear signals. \par

\begin{figure}[!h]
    \centering
    \includegraphics[width=120mm]{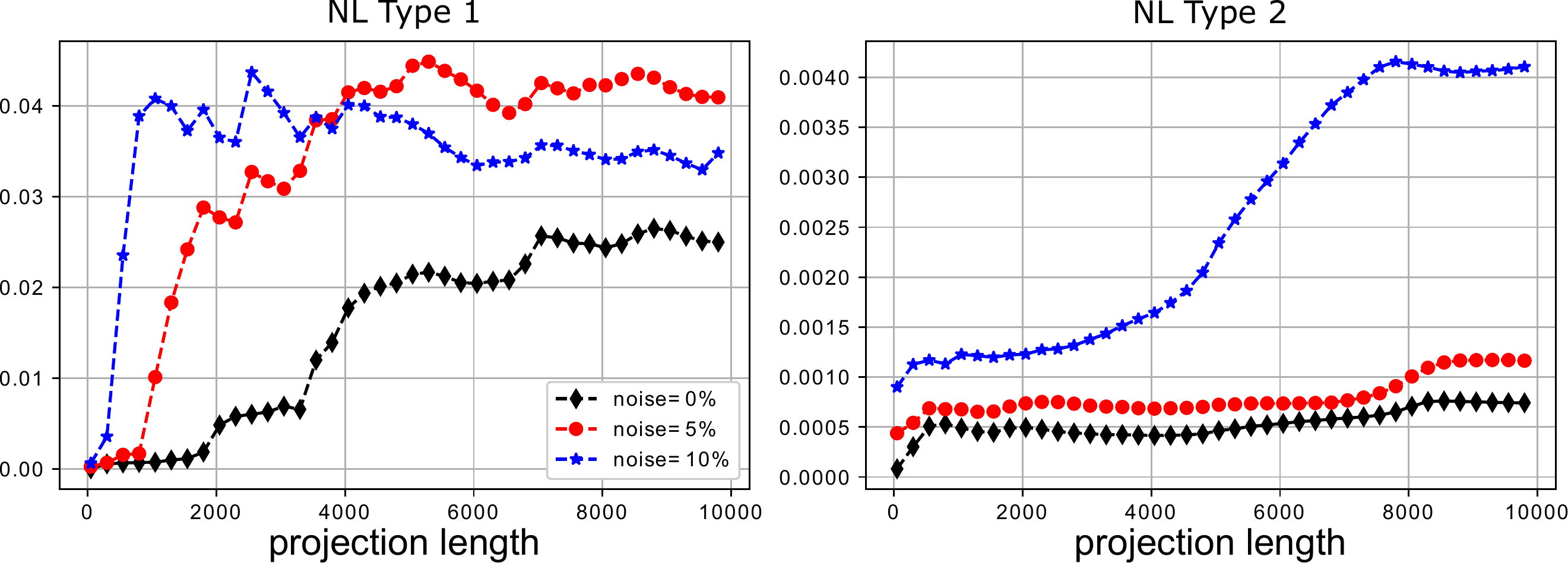}
    \caption{MSE error of the predicted signals vs. the length of projection. As expected, the error increases as the projection length is longer. However, in all cases after a rapid jump in the error at the beginning, the error flattens for longer projections. Notice as expected, noisier signals have higher MSE errors.}
    \label{fig:proj_error}
\end{figure}

In general, recurrent networks are prone to instability in longer trajectories \citep{salehinejad2017recent}. Error accumulation due to feeding the output of the network back is reported as the main source of this instability \citep{holden2017phase}. In this study, by physically constraining the network as well as utilizing projection loss for training, the model enjoys stability for longer trajectory predictions. Figure \ref{fig:proj_error} shows prediction errors for different noise levels with respect to different projection lengths. For the noiseless case, the mean squared error (MSE) gradually increases as the trajectory lengthens. However, the error is still very low for very long trajectories (i.e., $10,000$ one-step ahead predictions equivalent to 200 seconds). Interestingly, for two noisy cases, except for the lower range of trajectories, the error remains constant for longer trajectories. This implies that: (1) DynNet is quite stable regardless of the trajectory length; and (2) noisier data tends to discount the increasing error issue for longer trajectories. \par

\begin{figure}[!h]
    \centering
    \includegraphics[width=145mm]{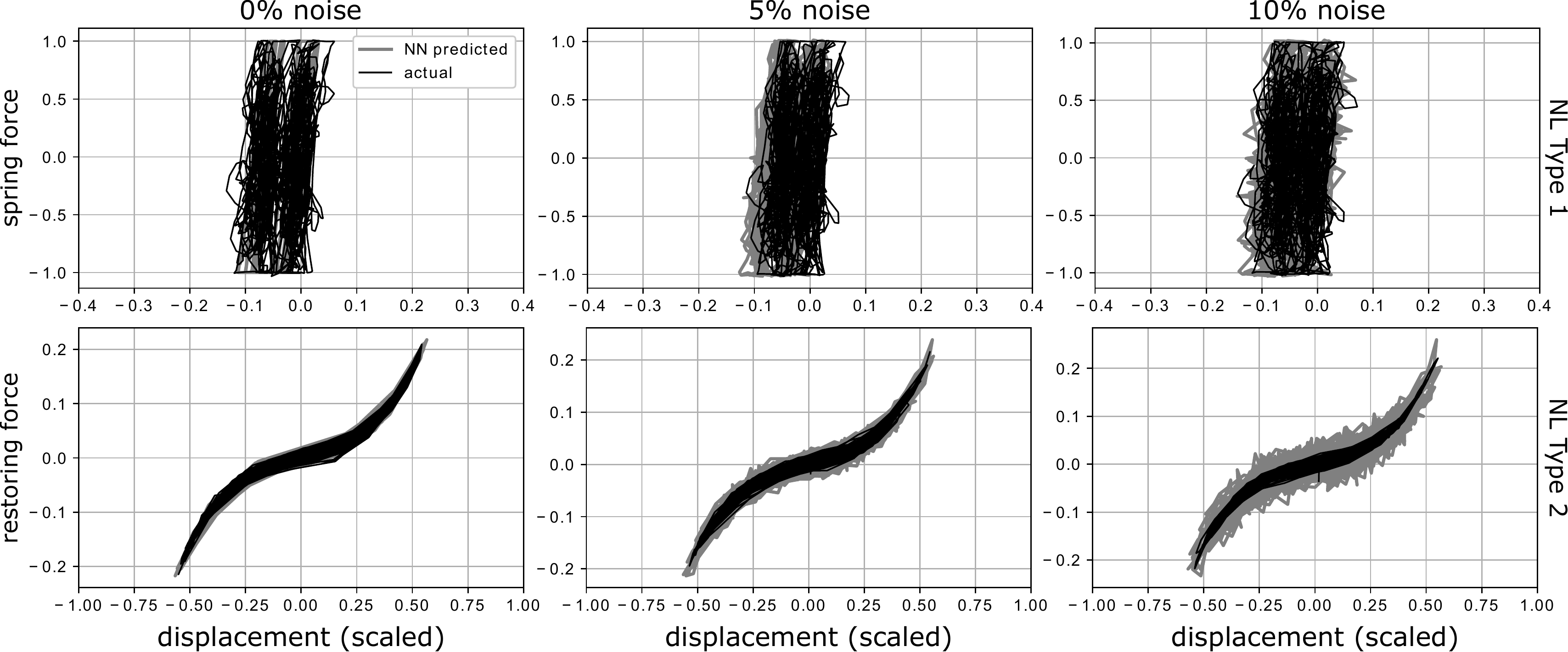}
    \caption{Hysteresis diagram in two nonlinearity cases at the first floor and for different noise levels. Both sets of results confirm the promising performance of the network in learning different nonlinear behaviors.}
    \label{fig:hyster}
\end{figure}

Finally, to verify the strength of DynNet in identifying the nonlinear behavior, hysteresis diagrams for a randomly picked signal and different noise levels are shown in Figure \ref{fig:hyster}. The DynNet estimated signals could very accurately capture the linear tangent of the spring force. In addition, the transition to nonlinear region is learned very accurately (normalized force values are exactly bounded within -1 to 1). The same level of accuracy is noticeable in all noise cases. \par

\begin{figure}[!h]
    \centering
    \includegraphics[width=100mm]{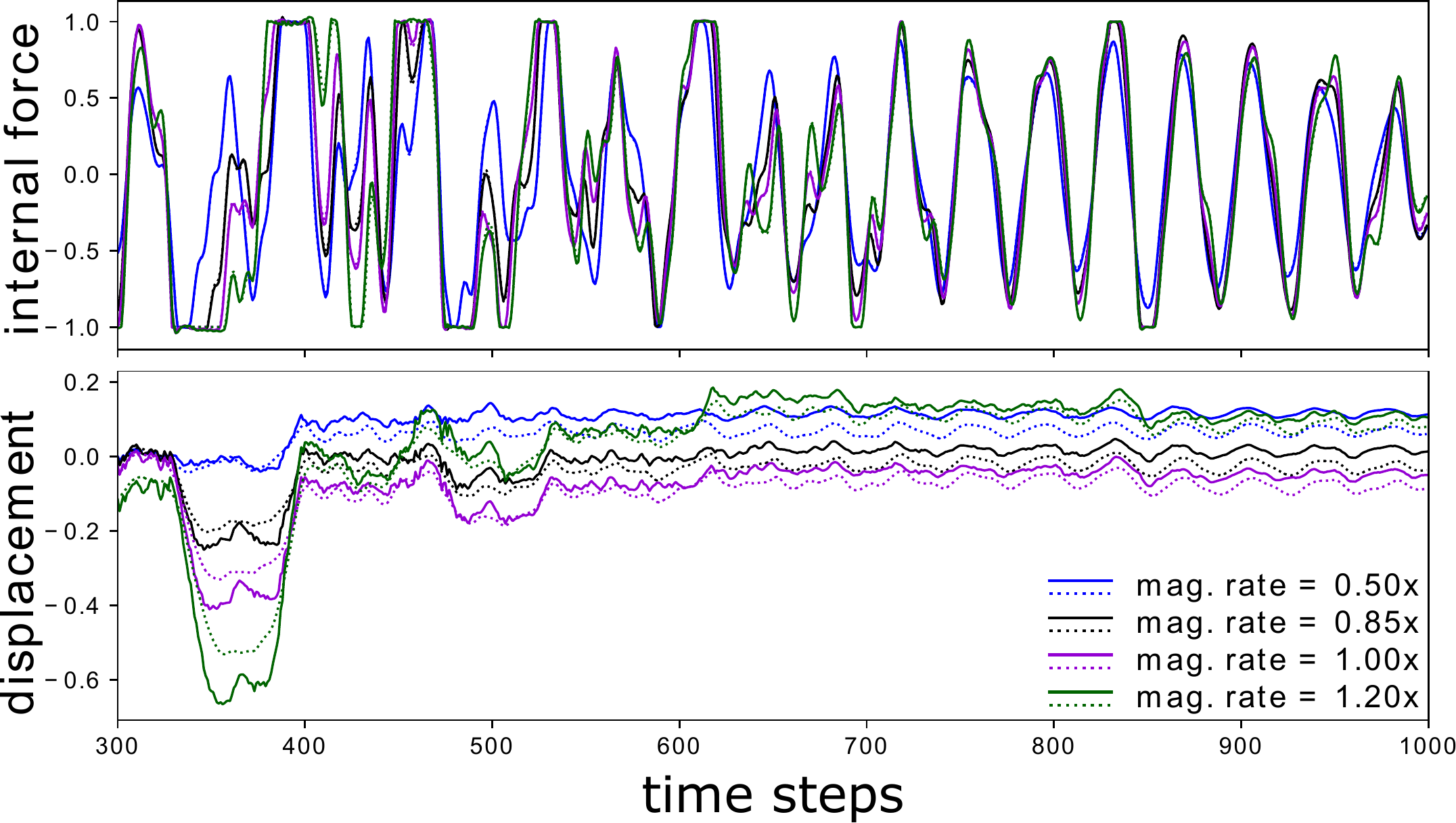}
    \caption{Response predictions for a ground motion with different magnitudes. This figure demonstrates the generalization of the trained NN model. Note that dashed lines show the actual responses from the numerical simulation. Despite strong nonlinear behavior, all four different magnitudes are predicted very accurately.}
    \label{fig:sigmag}
\end{figure}

To further investigate the scalability and generalization of the trained DynNet, the nonlinear responses of the structure subjected to different magnitudes of a selected earthquake ground motion are inferred and compared with the numerical solutions. Four levels of magnitude are considered in this analysis: 0.5x, 0.85x, 1.0x, and 1.2x (compared to the normalized ground motion). The results are presented in Figure \ref{fig:sigmag}. In this plot, dotted lines show exact simulation results while solid lines represent DynNet predictions. Results of internal forces and displacements for the $1^{st}$ DOF are shown for brevity. Internal forces are very accurately predicted in all four levels of magnitude of the ground motion. The accuracy is lower in the displacement predictions, however, the relative trends and lower amplitudes are carefully captured by DynNet. Note that the selected ground motion contains a strong shock-wave at $\sim380^{th}$ time step which causes severe nonlinear response and baseline shift (residual deformation) in the displacement predictions. The model, however, is still successful in following the exact variations of the building responses. 

\subsection{Case 2: Nonlinear Elastic Model (NL type 2)}

In the second case study, a 4-DOF structure with nonlinear elastic springs is studies. For the nonlinear springs, a $3^{rd}$ order polynomial behavior is introduced which can model a hardening after initial pseudo-linear phase (see Figure \ref{fig:forcedisp}). Due to the elasticity of the model, no residual displacements are expected here. Mechanical properties of the building are presented in Table \ref{tbl:nltype2}. In this table, M's and K's are defined as explained before. $k1$ and $k2$ are coefficients of the $3^{rd}$ order restoring force equation (Equation \ref{eq:functions}). The training process is identical to the previous case study. DynNet requires no pre-processing or special accommodation for different nonlinear models. The model is trained for the same number of iterations and epochs as the previous test case. Loss function variation over the iterations is shown in Figure \ref{fig:lossTrends}. Again, sudden drops in loss values are detected when the projection length of the custom loss increases. \par

\begin{table}[!h]
\centering

\caption{Mechanical properties for NL type 2.}
\begin{tabular}{l l l}
\toprule
\textit{Mechanical props.} & \textit{Values} & \textit{Units}\\
\midrule
M1                                                                & 0.340   & $kip.s^2/in$ \\ \hdashline  
M2/M1                                                             & 0.8    & -                           \\\hdashline  
M3/M1                                                             & 0.75   & -                           \\\hdashline  
M4/M1                                                             & 0.6    & -                           \\\hdashline  
K1                                                                & 100    & $kips/in$                     \\\hdashline  
K2/K1                                                             & 3/4    & -                           \\\hdashline  
K3/K1                                                             & 1/2    & -                           \\\hdashline  
K4/K1                                                             & 1/4    & -                           \\\hdashline  
k1                                                                & 1      & -                           \\\hdashline  
k2                                                                & 10     & $in^2$  \\
\bottomrule
\end{tabular}
\label{tbl:nltype2}
\end{table}

\begin{figure}[!h]
    \centering
    \includegraphics[width=120mm]{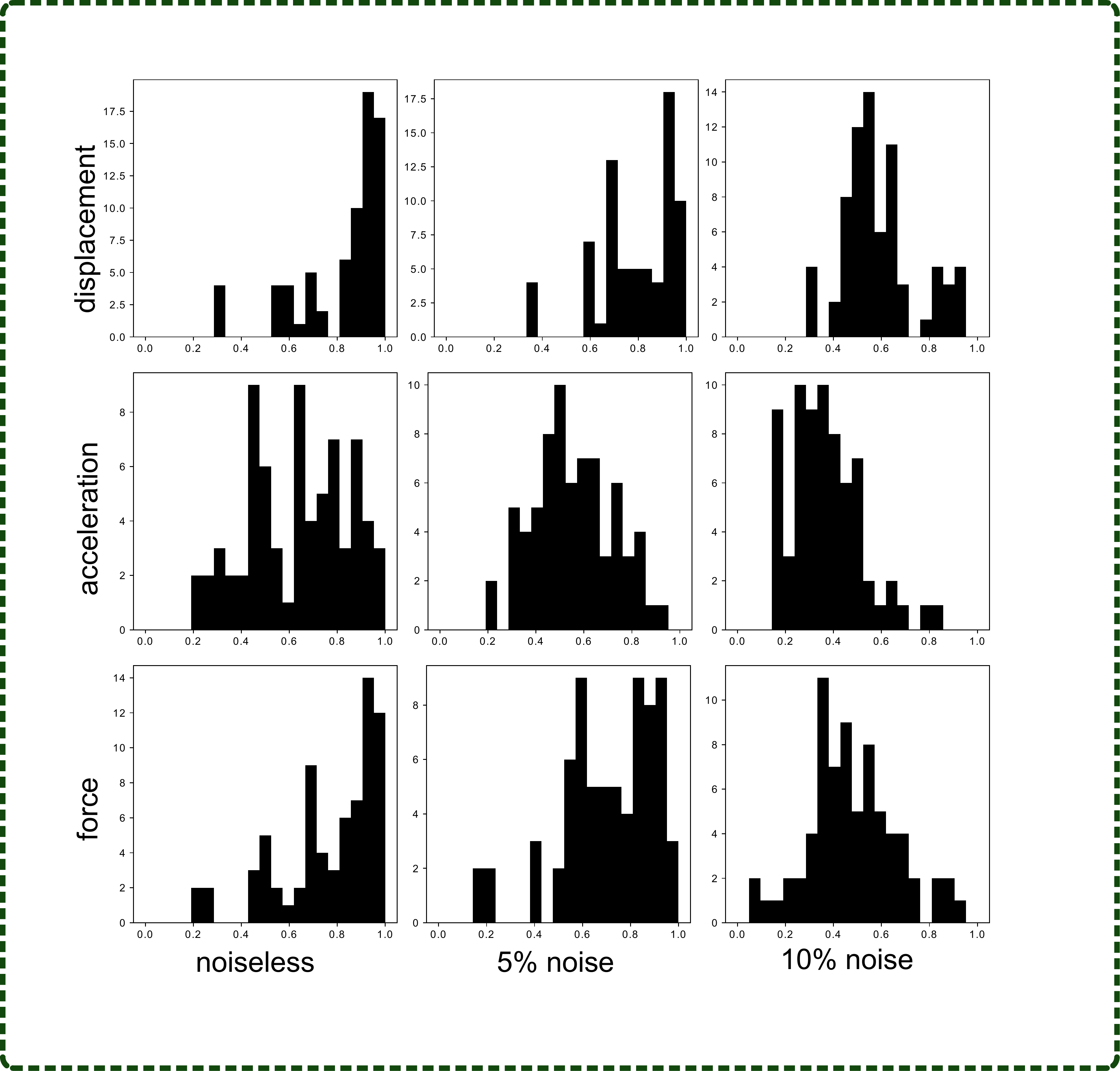}
    \caption{Pearson correlation coefficient histogram for predicted responses - nonlinearity type 2.}
    \label{fig:hist_NL2}
\end{figure}

The nonlinear response predictions for a randomly picked ground motion from testing data are presented in Figures \ref{fig:sig250} and \ref{fig:sig2000} (short and long trajectories, respectively). As before, DynNet shows a promising performance in nonlinear response predictions, both in time and frequency domains, regardless of the length of trajectory. To evaluate the predicting performance of the trained neural network on the entire testing data, PCC coefficients are calculated and the distributions are shown in Figure \ref{fig:hist_NL2}. Note that similar to the previous test case, three levels of measurement noise are considered for both training and evaluation phases of the network  
loss function. \par

In Figure \ref{fig:hist_NL2}, the general note is that the number of very high accuracy predictions (i.e., with PCC above 0.8) is not as high as the previous case, especially when measurement noise is introduced. However, for noiseless and $5\%$ additive noise cases, the results show high accuracy. Histogram of displacement and internal force response predictions show a unimodel distribution with the mode at PCC $\in [0.95,1.0]$. In terms prediction stability for longer trajectories, regression MSE error with respect to length of prediction trajectory is presented in Figure \ref{fig:proj_error}. Again, as observed in the PCC histograms, two lower noise cases show a steady trend of MSE loss as the trajectory length increases while the $10\%$ noise case is not as stable. Notice that the values of MSE errors generally are significantly lower in the NL type 2 (nonlinear elastic case) compared to the NL type 1 (elastic-perfectly plastic) while histograms show higher accuracy for prediction of the latter model. This observation is explained by the inelastic behavior of the NL type 1 model which can cause baseline shifts (i.e., residual deformations). We showed that DynNet is successful in capturing baseline variations, even though a small discrepancy causes much larger MSE errors for these response predictions. The baseline variations are not expected in the elastic model. \par

Finally, in order to validate the ability of the neural network to predict nonlinear elastic behavior of the spring forces, hysteresis diagrams are plotted in Figure \ref{fig:hyster}. The restoring force here includes both the elastic spring force and the damping force (i.e., $c\dot{x} + f(x)$ in Equation \ref{eq:eom}). According to these plots, DynNet predictions very accurately match with the simulation results. The $3^{rd}$ order behavior of the spring as well as the small energy dissipation area caused by the damper force is identified and correctly predicted. In higher noise levels, the prediction shows higher fluctuations around the exact plots which can be simply explained by the high level of noise.

\section{Conclusion}\label{sec:conclusion}

In this study, we proposed a data-driven approach for comprehensive prediction of nonlinear dynamic responses of multi degrees of freedom (DOF) systems using Neural Networks. In particular, inspired by common implicit dynamic analysis algorithms, DynNet block is designed as a one-step ahead response predictor. By repeatedly inferring the block, long response trajectories are predicted. Compared to the most advanced data-driven methods, DynNet has significantly smaller variable space, resulting less computational effort per iteration. Due to physics-based constraints of the proposed architecture, the network required more advanced optimizers for a smooth and efficient learning process. With this regard, trust-region approach using CG-Steihaug (TRCG) algorithm was implemented. In addition, for more efficient learning, a simple hardsampling technique as well as trajectory loss function was developed and implemented which resulted in faster learning of severely nonlinear transitions. \par

For verification, DynNet was tested in two nonlinear case studies: a four DOF shear building (1) with elastic perfectly plastic stiffness, and (2) with nonlinear elastic ($3^{rd}$ order) stiffness. For each test case, three levels of measurement noise were included to evaluate the noise propagation characteristics of the proposed network. The networks were trained using less than $30\%$ of the available data and evaluated using the remaining $70\%$. In both test cases, we showed that the network quite successfully was able to predict a complete set of nonlinear responses including displacement, velocity, acceleration, and internal force time histories at all DOFs given the applied ground motion only. The stability of the predictions for longer trajectories was analyzed and concluded that for the majority of cases, DynNet holds the error level stably as the trajectory length grows. In addition, using hysteresis diagrams, we showed that the performance of DynNet in capturing nonlinear behaviors of the springs is promising. \par

Data-driven function estimators are extremely popular in science and technology, however, in engineering applications due to the availability of accurate governing equations and numerical solutions, fully black-box function estimators are less accepted. This study tries to bridge the gap between black-box models and available exact solutions to create a fast learner function estimator. It is believed that DynNet can create a great potential for faster regional disaster sustainability and health monitoring analyses.

\section{Acknowledgments}

Research funding is partially provided by a grant from the U.S. Department of Transportation’s University Transportation Centers Program, and National Science Foundation through Grants CMMI-1351537, CCF-1618717, and CCF-1740796, and by a grant from the Commonwealth of Pennsylvania, Department of Community and Economic Development, through the Pennsylvania Infrastructure Technology Alliance (PITA).

\bibliographystyle{unsrtnat} 
\bibliography{references}

\end{document}